\documentclass{article}
\usepackage{amsmath,graphicx}

\usepackage{geometry}
\usepackage{amssymb}
\usepackage{array}
\usepackage{algorithm}
\usepackage{algorithmic}
\usepackage{url}
\usepackage{hyperref}
\usepackage{bbm}

\newcommand{\mx}[1]{\boldsymbol{\mathbf{#1}}}
\newcommand{\operator}[1]{\mathcal{#1}}
\newcommand{\R}{\mathbb{R}}

\newtheorem{theo}{Theorem}
\begin{document}

\title{Recovering Multiple Nonnegative Time Series From a Few Temporal Aggregates}

\author
 {Jiali~Mei$^{\star \dagger}$ \qquad Yohann~De~Castro$^{\dagger}$ \qquad Yannig~Goude$^{\star \dagger}$ \qquad Georges~H\'ebrail$^{\star}$ \\ \\
 $^{\star}$EDF Lab Paris-Saclay, 91120 Palaiseau, France. 
 \\ $^{\dagger}$LMO, Univ. Paris-Sud,
CNRS,  Universit\'e Paris-Saclay, 91405 Orsay, France.}

\maketitle

\begin{abstract}
Motivated by electricity consumption metering, we extend existing nonnegative matrix factorization (NMF) algorithms to use linear measurements as observations, instead of matrix entries.
The objective is to estimate multiple time series at a fine temporal scale from temporal aggregates measured on each individual series. 
Furthermore, our algorithm is extended to take into account individual autocorrelation to provide better estimation, using a recent convex relaxation of quadratically constrained quadratic program. 
Extensive experiments on synthetic and real-world electricity consumption datasets illustrate the effectiveness of our matrix recovery algorithms.
\end{abstract}

\section{Introduction}

In this paper, we propose to use a matrix recovery algorithm using nonnegative matrix factorization (NMF, \cite{lee_learning_1999}) on temporal aggregates on multiple time series. 
This is motivated by the use case of residential electricity metering. 
Residential utility consumption used to be recorded only once every few months. 
Nowadays, with smart meters, measurements may be recorded locally every minute or second but only transmitted as daily sums, due to data transmission, processing costs and/or privacy issues. 
Recent advances in matrix completion have made it clear that when a large number of individuals and features are involved, even partial data could be enough to recover much of lost information, thanks to the low-rank property: 
although the whole data matrix $\mx{V}^\ast \in \R^{T\times N}$ is only partially known, if 
$\mx{V}^\ast = \mx{W} \mx{H}$, where $\mx{W} \in\R^{T\times K}, \mx{H}\in\R^{K\times N}$, with $K$ much smaller than both $T$ and $N$, one might be able to recover $\mx{V}^\ast$ entirely.

Consider a nonnegative matrix $\mx{V}^\ast \in \R_+^{T\times N}$ of $N$ time series of $T$ periods. 
An entry of this matrix, $v^*_{t, n}$ represents, for example, the electricity consumption of Client $n$ for Period $t$. 
We consider the case where the observations consist of $D$ temporal aggregates, represented by a data vector $\mx{b} \equiv \operator{A}(\mx{V}^\ast) \in \R_+^D$, where $\operator{A}$ is a $D$-dimensional linear operator. 
To recover $\mx{V}^\ast$ from the measurements $\mx{b}$, we search a low-rank NMF of $\mx{V}^\ast$: 
$\mx{W} \mx{H} \simeq \mx{V}^\ast $, where $\mx{W} \in\R_+^{T\times K}, \mx{H}\in\R_+^{K\times N}$. 
The columns of $\mx{W}$ are $K$ nonnegative factors, interpreted as typical profiles of the $N$ time series, and the columns of $\mx{H}$ contain the weights of each individual. 
To estimate $\mx{W}$ and $\mx{H}$, we minimize a quadratic loss function under nonnegativity and data constraints:
 \begin{equation}
 \begin{aligned}\label{eq:mask_base}
 \min_{\mx{V} \text{, } \mx{W} \text{, } \mx{H}}\quad & \ell(\mx{V}, \mx{W}, \mx{H}) = \|\mx{V} - \mx{W}\mx{H} \|_F^2\\
 \text{s.t.}\quad & \mx{V}\geq \mx{0}, \quad\mx{W}\geq \mx{0}, \quad\mx{H}\geq \mx{0}, \quad\operator{A}(\mx{V}) = \mx{b},
 \end{aligned}
 \end{equation}
where $\mx{X} \geq \mx{0}$ (or $\mx{x} \geq \mx{0}$) means that the matrix $\mx{X}$ (or the vector $\mx{x}$) is element-wise nonnegative.

Our measurement operator $\operator{A}$ is a special instance of the trace regression model \cite{rohde_estimation_2011} which generalizes the matrix completion setting \cite{candes_exact_2009}. 
In matrix completion each measurement is exactly one entry. Various forms of linear measurements other than matrix completion have been considered for matrix recovery without nonnegativity \cite{recht_guaranteed_2010,candes_tight_2011,zuk_low-rank_2015}. 
To the best of our knowledge, most studies in NMF are either focused on full observation for dimension reduction \cite{gillis_why_2014,alquier2016oracle}, or on matrix completion \cite{gillis_low-rank_2011,xu_block_2013}. 
We extend classic alternating direction NMF algorithms to use linear measurements as observations. 
This extension is discussed in Section \ref{sec:base algo}.

In real-world applications, global information such as temporal autocorrelation could be available in addition to measurements. 
Previous approaches combining matrix factorization and autoregressive structure are often focused on obtaining factors that are more smooth and/or sparse, both in NMF \cite{chen_nonnegative_2005,fevotte_algorithms_2011,smaragdis_static_2014} and without nonnegativity \cite{udell_generalized_2014,yu_high-dimensional_2015}. 
Our objective is different from these studies: we try to further improve the matrix recovery by constraining temporal correlation on individual time series (not factors). 
We use a recent convex relaxation of quadratically constrained quadratic programs \cite{ben-tal_hidden_2013} to deduce an algorithm in this case.
This is discussed in Section \ref{sec:algo autocorrelation}. 
In Section~\ref{sec:experiments}, both algorithms are validated on synthetic and real electricity consumption datasets.

\section{Matrix of nonnegative time series}

\subsection{Iterative algorithm with simplex projection}\label{sec:base algo}

We represent temporal aggregation by a linear operator $\operator{A}$. 
For each $1\leq d \leq D$, the $d$-th measurement on $\mx{X}$, $\operator{A}(\mx{X})_d$, is the sum of several consecutive rows on a column of $\mx{X}$.
Each measurement covers a disjoint index set. Not all entries of $\mx{X}$ are necessarily involved in the measurements.

We use a block Gauss-Seidel algorithm (Algorithm~\ref{algo:linear mask}) to solve (\ref{eq:mask_base}). 
We alternate by minimizing $\ell(\mx{V}, \mx{W}, \mx{H})$ over $\mx{W}, \mx{H}$ or $\mx{V}$, keeping the other two matrices fixed. 
Methods from classical NMF problems are used to update $\mx{W}$ and $\mx{H}$ \cite{kim_algorithms_2014}. 
In our implementation, we use two variants that seem similarly efficient (more details in Section \ref{sec:experiments}): 
Hierarchical Alternating Least Squares (HALS, \cite{cichocki_hierarchical_2007}), 
and a matrix-base NMF solver with Nesterov-type acceleration (NeNMF, \cite{guan_nenmf:_2012}).

When $\mx{W}$ and $\mx{H}$ are fixed, the optimization problem on $\mx{V}$ is equivalent to $D$ simplex projection problems, one for each scalar measurement. 
For $1 \leq d \leq D$, we have
 \begin{equation}
 \begin{aligned}\label{eq:projection_mask}
 \min_{\mx{v}_{I_d}}\quad &\|\mx{v}_{I_d} - \sum_{t = t_0(d) + 1 }^{t_0(d) + h(d)} w_t\mx{h}_{n_d} \|^2\\
 \text{s.t.}\quad & \mx{v}_{I_d} \geq 0, \quad\mx{v}_{I_d}' \mx{1} = b_d.
 \end{aligned}
 \end{equation}
We use the simplex projection algorithm introduced by \cite{chen_projection_2011} to solve this subproblem efficiently. 
In Algorithm~\ref{algo:linear mask}, we note this step as the projection operator, $\operator{P}_A$, into the affine subspace $A\equiv \{\mx{X} \in \R_+^{T\times N} | \operator{A}(\mx{X}) = \mx{b} \}$. 
Projector $\operator{P}_A$ encodes the measurement data $\mx{b} = \operator{A}(\mx{V}^\ast)$.

\begin{algorithm}
\caption{Block coordinate descent for NMF from temporal aggregates}
\label{algo:linear mask}
\begin{algorithmic}
\REQUIRE $\mathcal{P}_A, 1\leq K \leq \min\{T,N\}$
\STATE Initialize $\mx{W}^0, \mx{H}^0 \geq 0, \mx{V}^0 = \mathcal{P}_A (\mx{W}^0 \mx{H}^0), i = 0$\;
\WHILE{Stopping criterion is not satisfied}
\STATE $\mx{W}^{i+1} = \text{Update}(\mx{W}^{i}, \mx{H}^{i}, \mx{V}^{i})$\;
\STATE  $\mx{H}^{i+1} = \text{Update}(\mx{W}^{i+1}, \mx{H}^{i}, \mx{V}^{i})$\;
\STATE  $\mx{V}^{i+1} = \mathcal{P}_A(\mx{W}^{i+1} \mx{H}^{i+1})$\;
\STATE  $i = i+1$\;
\ENDWHILE
\RETURN $\mx{V}^i \in A, 
 \mx{W}^i \in \mathbb{R}_+^{T\times K},
 \mx{H}^i \in \mathbb{R}_+^{K\times N}$
\end{algorithmic}
\end{algorithm}

We use a classical stopping criterion in the NMF literature based on Karush-Kuhn-Tucker (KKT) conditions on (\ref{eq:mask_base}) (\cite[Section 3.1.7]{gillis_why_2014}). 
We calculate
$
\mx{R}(\mx{W})_{i,j} = 
|(\mx{W} \mx{H} - \mx{V})\mx{H}')_{i,j}| \mathbbm{1}_{\mx{W}_{i,j} \neq 0}$, and $
\mx{R}(\mx{H})_{i,j} = 
|(\mx{W}' (\mx{W} \mx{H} - \mx{V}))_{i,j}|\mathbbm{1}_{\mx{H}_{i,j} \neq 0}.
$
The algorithm is stopped if $\|\mx{R}(\mx{W})\|_F^2 + \|\mx{R}(\mx{H}) \|_F^2 \leq \epsilon$, for a small threshold $\epsilon >0$.

Algorithm \ref{algo:linear mask} can be generalized to other types of measurement operators $\operator{A}$, as long as a projection into the affine subspace verifying the data constraint $\operator{A}(\mx{X}) = \mx{b}$ can be efficiently computed.

\subsection{From autocorrelation constraint to penalization\label{sec:algo autocorrelation}}

In addition to the measurements in $\mx{b}$, we have some prior knowledge on the temporal autocorrelation of the individuals. 
For $1 \leq n \leq N$, suppose that the lag-1 autocorrelation of Individual $n$'s time series is at least equal to a threshold $\rho_n$ (e.g. from historical data), that is, 
\begin{align}\label{eq:autocorrelation_constraint}
\sum_{t = 1}^{T-1} v_{t+1, n} v_{t, n} \geq \rho_n \sum_{t=1}^T v_{t,n}^2 .
\end{align}
Notice that with the lag matrix, $\mx{\Delta} = \begin{pmatrix}
0 & 1 & 0 & ... & 0 \\
0 & 0 & 1 & ... & 0 \\
0 & 0 & 0 & \ddots & \colon \\
\colon & \colon & \ddots & \ddots & 1 \\
0 & 0 & ... & 0 & 0 
\end{pmatrix}$, $\sum_{t = 1}^{T-1} v_{t+1, n} v_{t, n} = \mx{v}_n' \mx{\Delta} \mx{v}_n$. 
We define $\mx{\Delta}_{\rho} \equiv \mx{\Delta} + \mx{\Delta}' - 2\rho \mx{I}$, for a threshold $-1 \leq \rho \leq 1$.
Inequality (\ref{eq:autocorrelation_constraint}) is then equivalent to 
$
\mx{v}_n' \mx{\Delta}_{\rho_n} \mx{v}_n \geq 0.
$
To add an additional projection step in Algorithm \ref{algo:linear mask} would mean solving $N$ quadratically constrained quadratic programs (QCQP) of the form:
\begin{equation}
\begin{aligned}\label{eq:QCQP}
\min_{\mx{x}} \quad & ||\mx{x} - \mx{x}_0||^2 \\
\text{s.t. } \quad & \mx{x}' \mx{\Delta}_\rho \mx{x} \geq 0.
\end{aligned}
\end{equation}
Let $\mx{\Delta}_{\rho} = \mx{U}'\mx{D}_{\rho}\mx{U}$ be a diagonalization of $\mx{\Delta}_{\rho}$. 
The matrix $\mx{U}$ is orthogonal, because $\mx{\Delta}_{\rho}$ is symmetric. 
Let $\mx{\delta}_{\rho}$ be the diagonal entries of $\mx{D}_{\rho}$. 
In our case, $\delta_{\rho, t} = 2\cos(\frac{t}{T+1} \pi) - 2 \rho$, $\forall 1 \leq t \leq T$.
This means that (\ref{eq:QCQP}) is non-convex for most of $-1 \leq \rho \leq 1$. Nevertheless, the following theorem shows that its optimum can be computed relatively easily.
\begin{theo}\label{theo:QCQP}
 Suppose that $\mx{\delta}_{\rho}$ has at least one strictly positive component, and $\mx{U}\mx{x}_0$ has no zero component (generically satisfied). There exists $\lambda \geq 0$, so that $\mx{x}^\ast \equiv (\mx{I} - \lambda\mx{\Delta}_{\rho})^{-1} \mx{x}_0$ is the optimal solution of (\ref{eq:QCQP}).
\end{theo}

\textit{Proof.}
We follow \cite{ben-tal_hidden_2013} to get a convex relaxation of (\ref{eq:QCQP}). 
Notice that $\delta_{\rho, t}$ are all distinct (non-degenerate) eigenvalues. 
Define $\mx{z} = \mx{U}\mx{x}, \mx{z}_0 = \mx{U}\mx{x}_0, y_t = \frac{1} {2} z_t^2, \forall 1 \leq t \leq T$. 
Recall that $\exists \tau, 1\leq \tau \leq T$, $\delta_{\rho, \tau} >0$, and that ${\forall t, 1 \leq t \leq T}, {z_{0, t} \neq 0}$.

Problem (\ref{eq:QCQP}) is equivalent to the non-convex problem
\begin{equation}\label{eq:QCQP-equivalent}
\begin{aligned}
\min_{\mx{y}, \mx{z}} \quad & \mx{1}' \mx{y} - \mx{z}_0' \mx{z} \\
\text{s.t. } \quad & -\mx{\delta}_{\rho}' \mx{y} \leq 0, \quad \frac{1} {2} z_t^2 = y_t, \quad \forall 1 \leq t \leq T.
\end{aligned}
\end{equation}
Now consider the convex problem
\begin{equation}\label{eq:QCQP-relaxed}
\begin{aligned}
\min_{\mx{y}, \mx{z}} \quad & \mx{1}' \mx{y} - \mx{z}_0' \mx{z} \\
\text{s.t. } \quad & -\mx{\delta}_{\rho}' \mx{y} \leq 0, \quad \frac{1} {2} z_t^2 - y_t \leq 0, \quad \forall 1 \leq t \leq T.
\end{aligned}
\end{equation}
By \cite[Theorem 3]{ben-tal_hidden_2013}, if $(\mx{z}^\ast, \mx{y}^\ast)$ is an optimal solution of (\ref{eq:QCQP-relaxed}), and if $\frac{1}{2} (z_t^\ast)^2 = y_t^\ast$, $\forall 1\leq t \leq T$, then $(\mx{z}^\ast, \mx{y}^\ast)$ is also an optimal solution of (\ref{eq:QCQP-equivalent}), which makes $\mx{x}^{\ast} = \mx{U}'\mx{z}^\ast$ an optimal solution of (\ref{eq:QCQP}).

Problem (\ref{eq:QCQP-relaxed}) is convex, and it verifies the Slater condition: $\exists (\hat{\mx{y}}, \hat{\mx{z}}), -\mx{\delta}_{\rho}' \hat{\mx{y}} < 0, \frac{1} {2} \hat{z}_t^2 < \hat{y}_t, \forall 1 \leq t \leq T$. 
This is true, because $\delta_{\rho, \tau}>0$. 
We could choose $\hat{y}_\tau>0$ and small but strictly positive values for other components of $\hat{\mx{y}}$ so as to have $-\mx{\delta}_{\rho}' \hat{\mx{y}} < 0$, and $\hat{\mx{z}} = \mx{0}$. 
Thus, (\ref{eq:QCQP-relaxed}) always has an optimal solution, because the objective function is coercive over the constraint. 
This shows the existence of $(\mx{z}^\ast, \mx{y}^\ast)$.

Now we show that $\frac{1}{2} (z_t^\ast)^2 = y_t^\ast$, $\forall 1\leq t \leq T$.
The KKT conditions of (\ref{eq:QCQP-relaxed}) are verified by $(\mx{z}^\ast, \mx{y}^\ast)$. 
In particular, there is some dual variable $\lambda \geq 0, \mx{\mu} \in \R_+^T$ that verifies,
\begin{align}
 \mx{1} - \lambda \mx{\delta}_{\rho} - \mx{\mu} &= \mx{0}, \label{eq:derivative y}\\
 \lambda \mx{\delta}_{\rho}' \mx{y}^\ast & = 0, \label{eq:slackness y}\\
 -z_{0, t} + \mu_t z_t^\ast &= 0, \quad \forall 1 \leq t \leq T, \label{eq:derivative z}\\
 \mu_t (\frac{1} {2} (z_t^\ast)^2 - y_t^\ast) & = 0, \quad \forall 1 \leq t \leq T \label{eq:slackness z}.
\end{align}
Since $z_{0, t} \neq 0$, we have $\mu_{t} \neq 0, z^\ast_t = \frac{1}{\mu_t} z_{0, t}, \forall 1 \leq t \leq T$ by (\ref{eq:derivative z}). 
The values for $\mu_t = 1 - \lambda \delta_{\rho,t}$ can be deduced from (\ref{eq:derivative y}). 
By (\ref{eq:slackness z}), $y^\ast_t = \frac{1}{2}(z^{\ast}_t)^2, \forall 1 \leq t \leq T$. 
Therefore, $\bar{z_t} = z_t^\ast = \frac{z_{0,t}}{1 - \lambda \delta_{\rho, t}}, \forall 1 \leq t \leq T$. 
This shows that $\mx{x}^\ast = (\mx{I} - \lambda\mx{\Delta}_{\rho})^{-1} \mx{x}_0 = \mx{U}' \bar{\mx{z}}$ is an optimal solution for (\ref{eq:QCQP}). \hfill $\square$

Theorem \ref{theo:QCQP} shows that with a properly chosen $\lambda$, Problem (\ref{eq:QCQP}) can be easily solved. 
However, computing exactly $\lambda$ is computationally heavy. 
Therefore, we propose to replace the constraints by penalty terms in the objective function in (\ref{eq:mask_base}), which becomes 
 \begin{equation}
 \begin{aligned}\label{eq:mask_autocorrelation}
 \min_{\mx{V}, \mx{W}, \mx{H}}\quad &\|\mx{V} - \mx{W}\mx{H} \|_F^2 - 
   \lambda \sum_{n = 1}^N \mx{v}_n' \mx{\Delta}_{\rho_n} \mx{v}_n \\
 \text{s.t.}\quad & \mx{V}\geq 0, \quad\mx{W}\geq 0, \quad\mx{H}\geq 0, \quad\operator{A}(\mx{V}) = \mx{b},
 \end{aligned}
 \end{equation}
where $\lambda \geq 0$ is a single fixed regularization parameter. 

When $\mx{W}$ and $\mx{H}$ are fixed, the subproblem on $\mx{V}$ can be separated into $N$ constrained problems of the form,
\begin{equation}\label{eq:lagrangian of QCQP}
\begin{aligned}
\min_{\mx{x}} \quad & \|\mx{x} - \mx{x}_0\|^2 - \lambda \mx{x}'\mx{\Delta}_{\rho_n} \mx{x},\\
\text{s.t. } \quad & \mx{A}_n\mx{x} = \mx{c}_n,
\end{aligned}
\end{equation}
where $\mx{x}_0$ is the $n$-th column of $\mx{W}\mx{H}$, $\mx{c}_n$ is the observations on the $n$-th column, and $\mx{A}_n$ is a matrix which encodes the observation pattern over that column. 
The following theorem shows how to solve this problem.
\begin{theo}\label{theo:penalized constrained problem}
Define $\mx{Q}_n \equiv (\mx{I} - \lambda \mx{\Delta}_{\rho_n})^{-1}\mx{A}_n'(\mx{A}_n (\mx{I} - \lambda \mx{\Delta}_{\rho_n})^{-1} \mx{A}_n')^{-1}$.
Under the same conditions as in Theorem \ref{theo:QCQP}, if $\lambda < \frac{1}{\delta_{\rho, 1}}$, the minimizer for (\ref{eq:lagrangian of QCQP}) is $\mx{Q}_n\mx{c}_n + (\mx{I} - \mx{Q}_n\mx{A}_n)(\mx{I} - \lambda \mx{\Delta}_{\rho_n})^{-1} \mx{x}_0$.
\end{theo} 
\textit{Proof.}
Let $d_n$ be the dimension of $\mx{c}_n$. Define $I_C$ as the indicator function for the constraint of (\ref{eq:lagrangian of QCQP}), that is
$I_C(\mx{x}) = 
0 \text{ if } \mx{A}_n \mx{x} = \mx{c}_n, $ and $I_C(\mx{x}) = 
+\infty \text{ if } \mx{A}_n \mx{x} \neq \mx{c}_n.
$
Problem (\ref{eq:lagrangian of QCQP}) is then equivalent to 
\begin{align}\label{eq:indicator penalization}
\min_{\mx{x}} F(\mx{x}) \equiv \frac{1}{2}\|\mx{x} - \mx{x}_0\|^2 - \frac{1}{2}\lambda \mx{x}'\mx{\Delta}_{\rho_n} \mx{x} + I_C(\mx{x}).
\end{align}
The subgradient of (\ref{eq:indicator penalization}) is $\partial F(\mx{x}) = \{\mx{x} - \mx{x}_0 - \lambda \mx{\Delta}_{\rho_n} \mx{x} - \mx{A}_n' \mx{\epsilon} | \mx{\epsilon}\in \R^{d_n} \}$. 
When $\lambda < \frac{1}{\delta_{\rho, 1}}$, (\ref{eq:indicator penalization}) is convex. 
Therefore, $\mx{x}^\ast$ is a minimizer if and only if $0 \in \partial F(\mx{x})$, and $\mx{A}_n \mx{x}^\ast = \mx{c}_n$. 
That is,
$
\exists \mx{\epsilon}\in \R^{d_n}, \quad (\mx{I} - \lambda \mx{\Delta}_{\rho_n} ) \mx{x}^\ast - \mx{x}_0 - \mx{A}_n' \mx{\epsilon} = \mx{0}, \quad \mx{A}_n \mx{x}^\ast = \mx{c}_n.
$
The vector $\mx{\epsilon}$ thereby verifies $\mx{A}_n (\mx{I} - \lambda \mx{\Delta}_{\rho_n} )^{-1} (\mx{x}_0 + \mx{A}_n' \mx{\epsilon}) = \mx{c}_n$. 

The $d_n$-by-$d_n$ matrix $\mx{A}_n (\mx{I} - \lambda \mx{\Delta}_{\rho_n} )^{-1} \mx{A}_n'$ is invertible, because $d_n$ is smaller than $T$, and $\mx{A}_n$ is of full rank (because each measurement covers disjoint periods). 
Therefore,
\begin{align*}
 \mx{\epsilon} &= (\mx{A}_n (\mx{I} - \lambda \mx{\Delta}_{\rho_n} )^{-1} \mx{A}_n')^{-1} (\mx{c}_n - \mx{A}_n (\mx{I} - \lambda \mx{\Delta}_{\rho_n} )^{-1} \mx{x}_0),\\
 \mx{x}^\ast &= (\mx{I} - \lambda \mx{\Delta}_{\rho_n} )^{-1}(\mx{x}_0 + \mx{A_n}' \mx{\epsilon}) \\
 &= \mx{Q}_n\mx{c}_n + (\mx{I} - \mx{Q}_n\mx{A}_n)(\mx{I} - \lambda \mx{\Delta}_{\rho_n})^{-1} \mx{x}_0.\square
\end{align*}

Both $\mx{I} - \lambda \mx{\Delta}_{\rho_n}$ and $\mx{A}_n (\mx{I} - \lambda \mx{\Delta}_{\rho_n})^{-1} \mx{A}_n'$ are always invertible under the conditions of the theorem. 
The matrix inversion only needs to be done once for each individual. 
After computing $\mx{Q}_n\mx{c}_n$ and $(\mx{I} - \mx{Q}_n\mx{A}_n)(\mx{I} - \lambda \mx{\Delta}_{\rho_n})^{-1}$ for each $n$, we use Algorithm \ref{algo:autocorrel} to solve (\ref{eq:mask_autocorrelation}).

\begin{algorithm}
\caption{Block coordinate descent for NMF from temporal aggregates and autocorrelation penalty}
\label{algo:autocorrel}
\begin{algorithmic}
\REQUIRE $\rho_n, \mx{A}_n, \mx{Q}_n, \mx{Q}_n\mx{c}_n, \forall 1 \leq n \leq N,$ and $ 1\leq K \leq \min \{T,N \}$
\STATE Initialize $\mx{W}^0, \mx{H}^0 \geq 0, \mx{V}^0 = \mathcal{P}_A (\mx{W}^0 \mx{H}^0), i = 0$\;
\WHILE{Stopping criterion is not satisfied}
\STATE $\mx{W}^{i+1} = \text{Update}(\mx{W}^{i}, \mx{H}^{i}, \mx{V}^{i})$
\STATE  $\mx{H}^{i+1} = \text{Update}(\mx{W}^{i+1}, \mx{H}^{i}, \mx{V}^{i})$
\FORALL {$1 \leq n \leq N$}
\STATE $\mx{v}_n^{i + 1} = \mx{Q}_n\mx{c}_n + (\mx{I} - \mx{Q}_n\mx{A}_n)(\mx{I} - \lambda \mx{\Delta}_{\rho_n})^{-1} \mx{W}^{i+1} \mx{h}^{i+1}_n$
\ENDFOR
\STATE  $i = i+1$\;
\ENDWHILE
\RETURN $\mx{V}^i \in A, 
 \mx{W}^i \in \mathbb{R}_+^{T\times K},
 \mx{H}^i \in \mathbb{R}_+^{K\times N}$
\end{algorithmic}
\end{algorithm}

Convergence to a stationary point has been proved for past NMF solvers with the full observation or the matrix completion setting (\cite{guan_nenmf:_2012,kim_algorithms_2014}). 
Our algorithms have similar convergence property. 
Although the subproblems on $\mx{W}$ and $\mx{H}$ do not necessarily have unique optimum, the projection of $\mx{V}$ attains a unique minimizer. 
By \cite[Proposition 5]{grippo_convergence_2000}, the convergence to a stationary point is guaranteed.

An optimal value of $\lambda$ could be calculated. 
Substituting the values of $\mx{y}^\ast$ in (\ref{eq:slackness y}), shows that $\lambda$ is a root of the polynomial
$\sum_{t = 1}^T \delta_{\rho, t} \frac{z_{0, t}^2}{2 (1-\lambda \delta_{\rho, t})^2}$. 
It also verifies $ 0< \lambda < \frac{1}{\delta_{\rho, 1} } $, where $ \delta_{\rho, 1}  = 2\cos(\frac{1}{T+1} \pi) - 2\rho$ is the biggest eigenvalue of $\mx{\Delta}_{\rho}$. 
The root-finding is too expensive to do at every iteration. 
In our experimental studies, we chose $\lambda = \min(1, \frac{1}{2 \max_n \delta_{\rho_n, 1}})$ in the penalization when the constraint in (\ref{eq:QCQP}) is active, and $\lambda = 0$ (no penalization) when the constraint is verified by $\mx{x}_0$.

\section{Experimental results}\label{sec:experiments}

We used one synthetic dataset and three real-world electricity consumption datasets to evaluate our algorithms. 
In each dataset, the individual autocorrelation is calculated on historical data from the same datasets, excluded in matrix recovery. 
\begin{itemize}
\item \textbf{Synthetic data}: 120 random mixtures of 20 independent Gaussian processes with Matern covariance function (shifted to be nonnegative) are simulated over 150 periods ($T = 150, N = 120, K =20$). 
\item \textbf{French electricity consumption} (proprietary dataset of Enedis): daily consumption of 636 groups (medium-volume feeders) of around 1,500 consumers based near Lyon in France during 2012 ($T = 365, N = 636$).
\item \textbf{Portuguese electricity consumption}\cite{UCI_ML} daily consumption in kW of 370 clients during 2014 ($T = 365, N = 370$).
\item \textbf{Irish residential electricity consumption}\cite{Cer_a,Cer_b}. daily consumption of 154 groups of 5 consumers during 200 days in 2010 ($T = 200, N = 154$).
\end{itemize}

For each individual in a dataset, we generated observations by selecting a number of observation periods. 
The temporal aggregates are the sum of the time series between two consecutive observation periods. 
The observation periods are chosen in two possible ways: 
periodically (at fixed intervals, with the first observation period sampled at random), 
or uniformly at random. 
The fixed intervals for periodic observations are $p \in \{ 2,3,5,7,10,15,30 \}$. 
With random observations, we used sampling rates that are equivalent to these intervals. 
That is, the number of observations $D$ verifies $\frac{D} {TN} = \frac{1}{p} \in \{ 0.5,0.33,0.2,0.14,0.1,0.07,0.03 \}$.

We applied both algorithms on each dataset, with ranks $2 \leq K \leq 20$. 
When autocorrelation penalization is used, we chose $\lambda = \min(1, \frac{1}{2 \max_n \delta_{\rho_n, 1} })$. 
With a recovered matrix $\mx{V}$ obtained in an algorithm run, we compute the normalized error in Frobenius norm: $\|\mx{V} - \mx{V}^\ast\|_F / \|\mx{V}^\ast\|_F$.
Error rates for the best rank (smallest oracle error) is then reported. 
In practice, the choice of rank should be dictated by the interpretation, i.e. the number of profiles ($K$) needed in the application.

\begin{figure}
\centering
\includegraphics[width=1\linewidth]{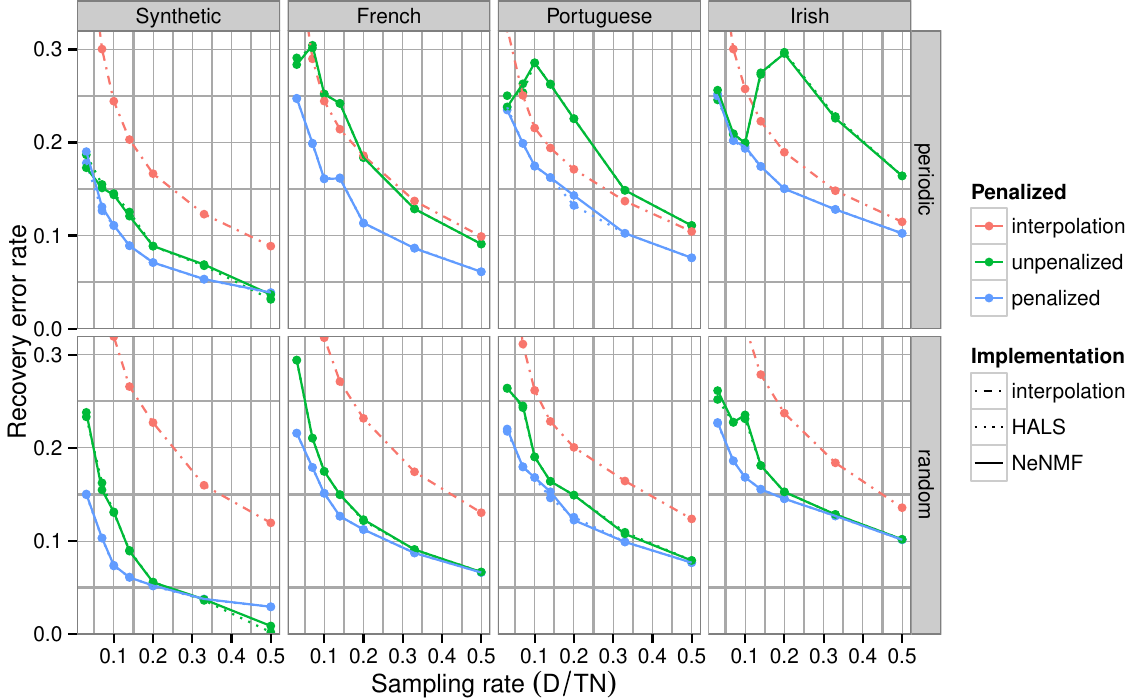}
\caption{Recovery error rate on synthetic and real-world datasets.}
\label{fig:schema}
\end{figure}

Each experiment (dataset, sampling scheme, sampling rate, unpenalized or penalized, implementation of update on $\mx{W}$ and $\mx{H}$) is run three times, and the average error rate is reported in Figure~\ref{fig:schema}. 
Both $\mx{W}$ or $\mx{H}$ update implementations (NeNMF with solid lines, HALS with dotted lines, almost identical) have equivalent error rates. 
As a comparison to our algorithms, we also provide the error rate of an interpolation method: 
temporal aggregates are distributed equally over the covered periods (red dot-dashed line). 
The figure is zoomed to show the error rates of the proposed algorithms. 
Much higher error rates for interpolation are sometimes not shown.

Proposed algorithms, whether unpenalized (green lines) or penalized (blue lines), out-performs the interpolation benchmark by large when the sampling rate is smaller than 10\%. 
With higher sampling rates, unpenalized recovery continues to works well on random observations (green lines in lower panels), but less so on real datasets with periodic observations (green lines in upper panels). 
Real electricity consumption has significant weekly periodicity, which is poorly captured by observations at similar periods. 
However, this shortcoming is more than compensated for by autocorrelation penalization (blue lines). 
Penalized recovery consistently outperforms interpolation with both observation schemes. 
This makes penalized recovery particularly useful for real-world applications, where it may be costly to change the sampling scheme from periodic to random (i.e. electricity metering).

\section{Perspectives}

We extended NMF to use temporal aggregates as observations, by adding a projection step into NMF algorithms. 
With appropriate projection algorithms, this approach could be further generalized to other types of data, such as disaggregating spatially aggregated data, or general linear measures.

When such information is available, we introduced a penalization on individual autocorrelation, which improves the recovery performance of the base algorithm. 
This component can be generalized to larger lags (with a matrix $\mx{\Delta}$ with 1's further off the diagonal), or multiple lags (by adding several lag matrices together). 
It is also possible to generalize this approach to other types of expert knowledge through additional constraints on $\mx{V}$.

\bibliographystyle{unsrt}
\bibliography{references,references_data}

\end{document}